\documentclass[11pt]{article}

\usepackage[preprint]{acl}

\usepackage{times}
\usepackage{latexsym}

\usepackage[T1]{fontenc}
\usepackage[utf8]{inputenc}

\usepackage{microtype}

\usepackage{inconsolata}

\usepackage{graphicx}

\usepackage{url}
\usepackage{amsmath}
\usepackage{booktabs}       % professional-quality tables
\usepackage{amsfonts}       % blackboard math symbols
\usepackage{nicefrac}       % compact symbols for 1/2, etc.
\usepackage{microtype}      % microtypography
\usepackage[table]{xcolor} % For row coloring
\definecolor{highlightgray}{gray}{0.92}
\usepackage{subcaption} % Required for creating subfigures

\usepackage{enumitem}
\usepackage{listings} % For code blocks

\usepackage[most]{tcolorbox}

\definecolor{promptHeader}{RGB}{60, 60, 60} % Dark gray for header
\definecolor{promptBody}{RGB}{240, 240, 240}   % Light gray for background

\definecolor{codegreen}{rgb}{0,0.6,0}
\definecolor{codegray}{rgb}{0.5,0.5,0.5}
\definecolor{codepurple}{rgb}{0.58,0,0.82}
\definecolor{backcolour}{rgb}{0.95,0.95,0.92}

\setlist[itemize]{noitemsep,leftmargin=*,topsep=0pt}
\setlist[enumerate]{noitemsep,leftmargin=*,topsep=0pt}

\title{TIM-PRM: Verifying multimodal reasoning with Tool-Integrated PRM}

\author{%
  Peng Kuang\textsuperscript{1,2}, Xiangxiang Wang\textsuperscript{2}, Wenchao Liu\textsuperscript{2}, Jian Dong\textsuperscript{2}, Kaidi Xu\textsuperscript{3} \\
  \textsuperscript{1}Zhejiang University, \textsuperscript{2}iFLYTEK AI Research Institute, \textsuperscript{3}City University of Hong Kong, \\ 
  \texttt{pengkuang@zju.edu.cn, \{xxwang56,wcliu4,jiandong3\}@iflytek.com}, \\
  kaidixu@cityu.edu.hk
}

\begin{document}
\maketitle
\begin{abstract}
Multimodal Large Language Models (MLLMs) have achieved impressive performances in mathematical reasoning, yet they remain vulnerable to visual hallucinations and logical inconsistencies that standard outcome-based supervision fails to mitigate. While Process Reward Models (PRMs) promise step-by-step verification, current approaches typically operate as scalar scorers or generative critics that suffer from sycophancy, blindly validating the flawed hypotheses rather than grounding them in visual reality. To bridge this gap, we introduce TIM-PRM (Tool-Integrated Multimodal PRM), a novel agentic framework that transforms verification from a passive classification task into an active, tool-augmented investigation. TIM-PRM is trained to explicitly plan verification strategies and utilizes a mechanism of Independent Question Asking to query evidence via external tools, effectively decoupling verification from the reasoning context to eliminate confirmation bias. We instantiate this method by curating a high-quality dataset of tool-integrated verification trajectories. Extensive experiments on VisualProcessBench demonstrate that our 8B parameter model surpasses existing open-source multimodal PRMs, significantly outperforming much larger models like Qwen2.5-72B and InternVL-78B, while offering interpretable insights into the verification process.
\end{abstract}

\section{Introduction}

The rapid advancement of Multimodal Large Language Models (MLLMs) has unlocked remarkable capabilities in solving complex reasoning tasks, ranging from geometry \cite{wang_measuring_2024,qiao_we-math_2025,zou_dynamath_2024,zhang_mathverse_2024} and scientific \cite{yue_mmmu_2024} analysis to visual logic puzzles. However, as these models are increasingly deployed in high-stakes domains, ensuring the reliability and faithfulness of their reasoning processes becomes paramount \cite{lightman_lets_2023,wang_visualprm_2025}. While standard alignment techniques, such as Reinforcement Learning from Human Feedback (RLHF) \cite{10.5555/3600270.3602281} with outcome supervision, have improved general response quality, they treat the reasoning chain as a black box. This outcome-focused paradigm often fails to penalize "false positives", where flawed reasoning serendipitously leads to the correct answer, thereby reinforcing hallucinated logic and diminishing the interpretability of the model's decision-making process.

To address these limitations, recent research has pivoted toward Process Supervision, where a Process Reward Model (PRM) \citep{lightman2024lets,li2025process,zheng2024processbenchidentifyingprocesserrors} evaluates intermediate steps of a solution. In the multimodal domain, existing approaches typically fall into two categories: discriminative PRMs that output scalar probabilities~\cite{wang_visualprm_2025, luo_ursa_2025}, and generative verifiers that produce linguistic critiques~\cite{noauthor_vrprm_2025,zhang_gm-prm_2025}. Despite their promise, both paradigms suffer from certain deficiencies. Scalar PRMs provide "black-box verification," offering no explanation for their scores and struggling to distinguish between subtle visual hallucinations and valid logical leaps. Conversely, generative verifiers often rely solely on the MLLM's internal parametric knowledge, making them prone to sycophancy: the tendency to blindly agree with the context of the reasoning step rather than critically inspecting the evidence \cite{liu_unveiling_2025}. Consequently, when a reasoning step asserts a false visual fact (e.g., "The graph is a parabola"), a standard verifier is biased to accept this premise, ignoring the error rather than identifying it. This raises the challenge of how to decouple verification from textual reasoning to create a verifier that is not only accurate but also explicitly grounded and resistant to confirmation bias.

\begin{figure*}[htbp]
    \centering
    \includegraphics[width=0.9\textwidth]{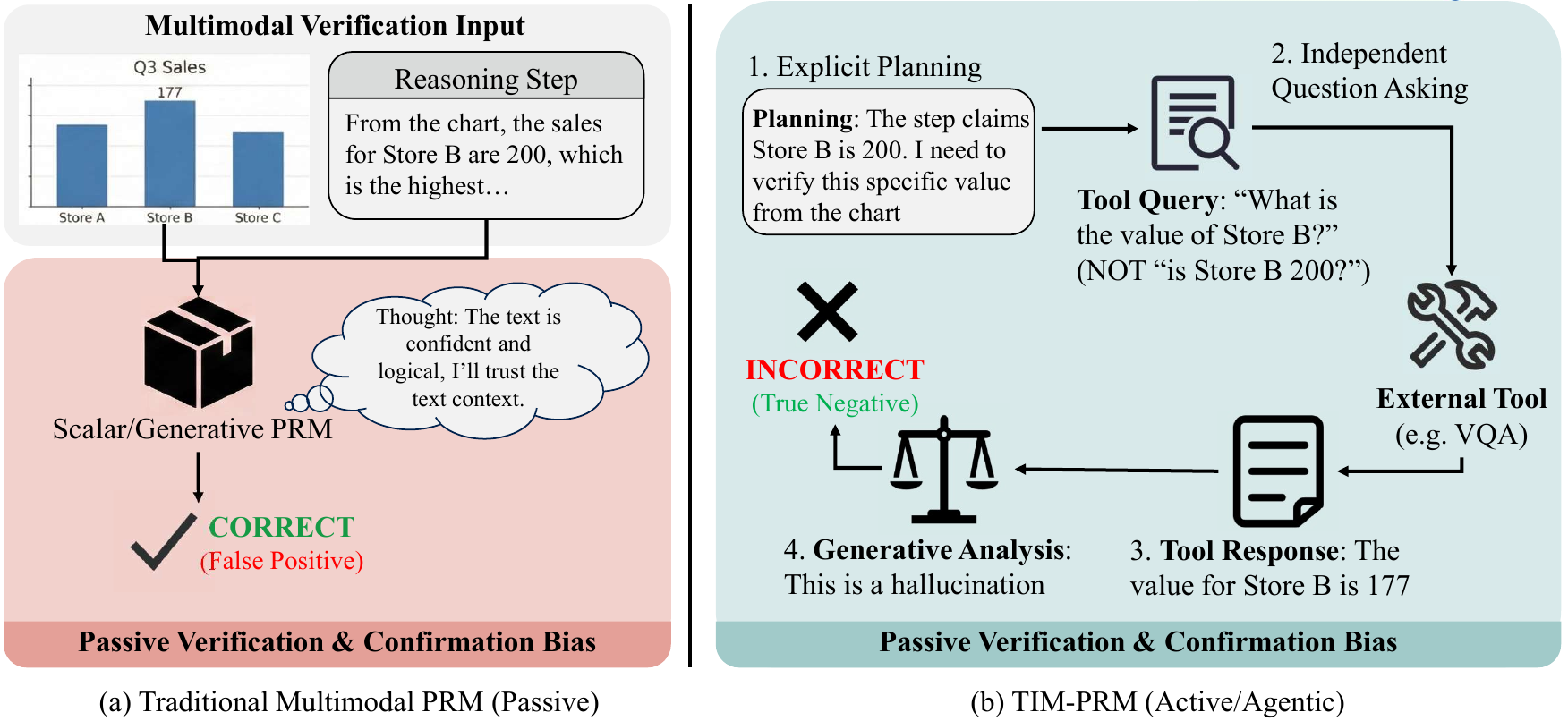}
    
    \caption{\label{fig:framework}Comparison between traditional passive verification and TIM-PRM. (a) Existing PRMs act as passive scorers, suffering from sycophancy where they blindly validate hallucinated claims due to reliance on text context. (b) TIM-PRM transforms verification into an agentic task, employing explicit planning, independent tool querying to decouple perception from reasoning, and grounded analysis to detect visual hallucinations.}
\end{figure*}

In this paper, we propose TIM-PRM (Tool-Integrated Multimodal PRM), a novel agentic framework that transforms verification from a passive scoring task into an active, tool-augmented investigation. Unlike traditional models that rely on ambiguous intuition, TIM-PRM is trained to explicitly \textit{plan} its verification strategy, determine what evidence is missing or considered uncertain by the model, and actively query the environment using external tools. To implement a flexible verification tool, we propose the mechanism of Independent Question Asking. Instead of verifying a hypothesis directly, which often triggers sycophancy, our model uses tools to ask open-ended questions about the image (e.g., "What is the shape of the graph?") before comparing the tool's response with the reasoning step. This extracts facts to be verified from the reasoning context, ensuring that the verdict is grounded in a reliable tool response rather than parametric hallucinations.

We instantiate TIM-PRM by curating a high-quality training dataset of tool-integrated verification trajectories, synthesized by a strong teacher model and filtered via Monte Carlo Tree Search (MCTS) consistency checks. We further introduce a sample upweighting strategy to counteract the inherent class imbalance in reasoning data, forcing the model to focus on the critical task of error detection. We evaluate our method on the challenging VisualProcessBench, covering diverse domains from MMMU to MathVision. Empirical results demonstrate that our 8B parameter model significantly outperforms state-of-the-art open-source models (including Qwen2.5-72B and InternVL-78B) and achieves competitive performance with proprietary giants like GPT-4o.

Our contributions are summarized as follows:
\begin{itemize}
    \item We identify key failure modes in existing multimodal verification, highlighting the issues of perception hallucinations in current PRMs.
    \item We introduce TIM-PRM, an agentic verification framework that integrates explicit planning and tool use to ground reasoning in external feedback.
    \item We propose a novel Independent Question Asking strategy that mitigates confirmation bias by decoupling visual perception from the reasoning hypothesis.
    \item Extensive experiments demonstrate that TIM-PRM surpasses open-source multimodal PRMs, particularly excelling in the First Incorrect Step Identification (FISI) metric, where it surpasses traditional scalar PRMs by a wide margin.
\end{itemize}

\section{Related works}

% \textbf{Multimodal Reward Modeling and Alignment.}
% Reward modeling serves as the cornerstone of aligning Multimodal Large Language Models (MLLMs) with human preferences and ensuring the correctness of generated outputs. Early approaches primarily focused on \textit{outcome supervision}, where a reward model evaluates the final response quality. 
% Works like InternLM-XComposer2.5-Reward \cite{zang_internlm-xcomposer25-reward_2025-1} and Skywork-VL Reward \cite{wang_skywork-vl_2025}
% pioneered this direction by training discriminative reward models on preference pairs to guide Reinforcement Learning from Human Feedback (RLHF) \cite{10.5555/3600270.3602281}.
% However, outcome-based supervision treats the reasoning process as a black box. While effective for general alignment, these methods often fail to locate intermediate logical errors or hallucinations in complex mathematical reasoning tasks, where a correct final answer can occasionally result from flawed reasoning (false positives).

\textbf{Process Supervision for Multimodal Reasoning.}
To address the opacity of outcome supervision, recent research has pivoted toward \textit{process supervision}, which evaluates reasoning trajectories step-by-step. This paradigm has shown significant promise in text-based math reasoning and is now being adapted for multimodal contexts.
VisualPRM~\cite{wang_visualprm_2025} and URSA~\cite{luo_ursa_2025} introduced discriminative Process Reward Models (PRMs) trained via Monte Carlo Tree Search (MCTS) derived labels. These models assign a scalar probability score to each step, guiding inference-time search algorithms like Best-of-N or Tree Search. Similarly, Athena~\cite{wang_athena_2025} demonstrated that data-efficient PRMs could be trained by leveraging error-injection techniques.
Despite their success, discriminative PRMs suffer from the "black-box verification" problem: they output a score without an explanation. As noted in our analysis (Section \ref{sec:failure_mode}), these models struggle with fine-grained visual grounding errors and often exhibit biases toward simple question replication or outcome-based heuristics. They lack the mechanism to explicitly "look" at the image to verify specific claims.

\textbf{Generative Verification and Tool Use.}
Moving beyond scalar scoring, \textit{generative verifiers} leverage the linguistic capabilities of LLMs to critique reasoning. In the text domain, methods have explored using LLMs to generate natural language feedback or self-corrections. 
In the multimodal domain, works like MM-RLHF~\cite{zhang_mm-rlhf_2025}, LLaVA-Critic~\cite{xiong_llava-critic_2025}, and R1-Reward~\cite{zhang_r1-reward_2025} pioneered in proposing generative reward models. More recently, VRPRM~\cite{noauthor_vrprm_2025} and GM-PRM~\cite{zhang_gm-prm_2025} have begun to explore generative process rewards, enabling PRMs to output reasoning traces alongside verification scores for each reasoning step.
However, current generative multimodal verifiers remain constrained by the internal parametric knowledge of the model. They are prone to sycophancy, often agreeing with the solution step's hallucinated content rather than challenging it. 
Our work distinguishes itself by integrating {explicit tool use} into the verification process. Unlike VRPRM or GM-PRM, which rely solely on internal chain-of-thought, TIM-PRM actively queries the visual environment through independent question-asking. This agentic approach decouples perception from reasoning, mitigating the hallucination propagation found in previous generative verifiers.

\section{Revisiting Verification of Multimodal Reasoning}

To understand the limitations of current approaches in verifing multimodal reasoning, we first conduct a preliminary evaluation of existing open-source multimodal PRMs. Through a fine-grained inspection of failure cases, we identify specific challenges related to visual perception and the inherent biases of probability-based reward modeling.

\subsection{Failure Modes in Visual Reasoning Verification}
\label{sec:failure_mode}

To investigate the root causes of verification failure, we conducted a qualitative error analysis. We randomly sampled 50 instances from VisualProcessBench where the VisualPRM model predicted the step correctness incorrectly. Through careful manual inspection, we categorized these failures into four distinct modes:

\begin{itemize}
    \item \textbf{Perception Error:} The step contains a hallucinatory description of the visual input (e.g., misidentifying a shape or reading a graph incorrectly). The PRM failed to flag this mismatch, indicating a lack of fine-grained visual grounding.
    
    \item \textbf{Knowledge Error:} The step relies on incorrect domain knowledge or formulas. Similar to perception errors, these went undetected, highlighting the model's limitations in internal knowledge retrieval.
    
    \item \textbf{First Step Bias:} The first step of a solution often simply restates or paraphrases the problem statement. While semantically neutral or correct, MCTS-based PRMs frequently assign these steps low scores. This artifact arises probably because on hard questions, even the first steps are assigned low scores by MCTS, despite being a valid reasoning step.
    
    \item \textbf{Final Step Bias:} Conversely, we observed cases where, in a correct trajectory, all previous steps are assigned with high scores, yet the final step is assigned with a low score, despite being logically sound and arriving at the correct final answer. This is probably due to the outcome-focused nature of MCTS, which we will elaborate on in the following subsection.
\end{itemize}

\begin{table}[h]
    \centering
    \caption{Distribution of failure modes in error analysis. The categories are aggregated from a manual inspection of verification failures. Note: Percentages are calculated based on the valid subset of steps ($N=57$), excluding incorrect labels and ground truth errors.}
    \label{tab:error_modes}
    \begin{tabular}{lrr}
        \toprule
        \textbf{Failure Mode} & \textbf{Count} & \textbf{Percentage} \\
        \midrule
        Perception Error & 20 & 35.1\% \\
        First Step Bias: & 19 & 33.3\% \\
        Knowledge Error & 8 & 14.0\% \\
        Final Step Bias & 6 & 10.5\% \\
        Others & 4 & 7.0\% \\
        \midrule
        \textbf{Total} & \textbf{57} & \textbf{100.0\%} \\
        \bottomrule
    \end{tabular}
\end{table}

\subsection{Analysis of Failure Modes}

The \textit{Perception} and \textit{Knowledge} errors underscore the fundamental challenge in verifying multimodal reasoning: the verifier is constrained by its own limited internal perception and parameters. This suggests a critical need for external tool use to ground verification in verifiable facts.

In contrast, the \textit{First Step Bias} and \textit{Final Step Bias} reveal flaws in the standard training data construction via MCTS. MCTS labels are derived from the rollout probability of reaching a correct solution, rather than the local semantic correctness of the step itself. Consequently, difficult problems may yield low scores for valid initial steps (due to low overall success rates), while outcome-focused labeling ignores logical gaps. 

These observations motivate our proposed approach: a verification method that utilizes explicit reasoning and external tools. By shifting from implicit probability estimation to explicit semantic analysis, we aim to constrain verification to the step's content rather than its rollout statistics.

\section{Methodology}

In this section, we introduce our \textbf{Tool-Integrated Multimodal PRM (TIM-PRM)}. Our method transitions from a scalar-based verifier to a generative, agentic framework capable of planning, querying visual data, and deducing correctness explicitly.

\subsection{Preliminary}

Let $\mathcal{Q}$ denote a multimodal math problem containing text and an image $\mathcal{I}$. A solution consists of a sequence of discrete steps $S = \{s_1, s_2, \dots, s_T\}$. The goal of a Process Reward Model (PRM) is to assign a verification label $v_t \in \{\text{Correct, Neutral, Incorrect}\}$ to each step $s_t$, conditioned on the problem and the history of previous steps $s_{<t}$. Unlike standard PRMs that approximate $P(v_t | \mathcal{Q}, \mathcal{I}, s_{\le t})$, our TIM-PRM generates a reasoning trajectory $\tau_t$ that includes tool execution traces before concluding the label $v_t$.

\subsection{Step-wise Verification with Tool Integration}

We construct our training data by synthesizing high-quality verification trajectories. Let $\pi_{\text{teacher}}$ denote a strong MLLM (e.g., Qwen3-VL-30B-A3B-Instruct) acting as the verifier. For a given step $s_t$ and history $s_{<t}$, the model generates a verification trajectory $\tau_t$ to maximize the probability $P(\tau_t, v_t | \mathcal{Q}, \mathcal{I}, s_{\le t})$.

We formalize the trajectory $\tau_t$ as a sequence of four distinct components: $\tau_t = (z_{\text{plan}}, z_{\text{call}}, z_{\text{resp}}, z_{\text{ana}})$, which leads to the final verdict $v_t$. The generation process is structured as follows:

\begin{itemize}
    \item \textbf{Explicit Planning ($z_{\text{plan}}$):} The model first generates a planning thought $z_{\text{plan}}$ to analyze $s_t$ given $\mathcal{I}$ and $s_{<t}$. This component identifies whether visual perception, external knowledge, or logical checking is required, effectively determining the verification policy $\pi(z_{\text{plan}} | \mathcal{Q}, \mathcal{I}, s_{\le t})$. This part is wrapped in the \texttt{<planning>} tag.
    \item \textbf{Tool Invocation ($z_{\text{call}}$):} Conditional on the plan, the model generates a structured tool call $z_{\text{call}}$. This separates the intent to verify from the execution, wrapped in the \texttt{<tool\_call>} tag.
    \item \textbf{Tool Response ($z_{\text{resp}}$):} The external environment executes the call in $z_{\text{call}}$ on image $\mathcal{I}$ and returns $z_{\text{resp}}$. This step injects external feedback into the trajectory, updating the context for the subsequent analysis. This part is wrapped in the \texttt{<tool>} tag.
    \item \textbf{Generative Analysis ($z_{\text{ana}}$):} The model synthesizes the tool feedback $z_{\text{resp}}$ (or internal knowledge) to generate a rationale $z_{\text{ana}}$. This ensures the final verdict is grounded in specific evidence, approximating $P(z_{\text{ana}} | \mathcal{Q}, \mathcal{I}, s_{\le t}, z_{\text{plan}}, z_{\text{call}}, z_{\text{resp}})$. This part is wrapped in the \texttt{<analyze>} tag.
    \item \textbf{Final Verdict ($v_t$):} Finally, the model concludes with a discrete judgment $v_t \in \{\text{Correct, Neutral, Incorrect}\}$ based on the generated chain of thought $\tau_t$. This part is wrapped in the \texttt{<verify>} tag.
\end{itemize}

To facilitate this, we format the solution steps into paragraphs and inject a system prompt defining the role of a "math teacher" and the available tools. Please refer to \textbf{Appendix \ref{app:prompt_structure}} for the specific prompt structure and data formatting details.

\textbf{Tool Execution Loop:}
The generation is an interactive process modeled as a Markov Decision Process (MDP) where the tool acts as the environment. We utilize a stop token to pause inference when $z_{\text{call}}$ is generated. The external tool is executed to obtain $z_{\text{resp}}$, which is appended to the context. The model then resumes to generate $z_{\text{ana}}$ and $v_t$.

\textbf{Data Filtering:}
To ensure the quality of the training data $\mathcal{D}$, we apply a rigorous filtering mechanism. We retain only those tuples $(\mathcal{Q}, \mathcal{I}, S, \mathcal{T}, \mathcal{V})$—where $\mathcal{T}=\{\tau_1 \dots \tau_T\}$ and $\mathcal{V}=\{v_1 \dots v_T\}$—that strictly follow the format constraints and where the verification labels $\mathcal{V}$ align with high-confidence MCTS search results.

\subsection{Debiasing Sycophantic Verification via Independent Question Asking}

% A significant challenge in verification is the tendency of MLLMs to be "sycophantic"—they often agree with the hypothesis presented in the step rather than critically evaluating it. For example, if a step claims "The graph is a parabola," a standard verifier is biased toward confirming this statement simply because it appears in the context.

% To address this, we design a specific tool: \texttt{ask\_questions}. This tool is motivated by the need to force the verification process to be inquisitive rather than confirmatory. Instead of validating a claim directly (e.g., "Is the graph a parabola?"), the model is encouraged to ask open-ended questions about the image (e.g., "What is the shape of the graph?") to gather independent facts. This strategy is inspired by recent findings \cite{liu_unveiling_2025} that MLLMs are more accurate and less biased when answering open-ended queries than when verifying true/false hypotheses.

A significant challenge in approximating $P(v_t | \mathcal{Q}, \mathcal{I}, s_{\le t})$ is the tendency of MLLMs to be "sycophantic"—where the probability mass of $v_t=\text{Correct}$ is artificially inflated by the presence of false premises in $s_t$. For instance, if $s_t$ claims "The graph is a parabola," the likelihood of the model validating this claim is biased by the textual context, regardless of $\mathcal{I}$.

To mitigate this, we introduce the mechanism of \textbf{Independent Question Asking} within the tool call component $z_{\text{call}}$. Instead of verifying a hypothesis $h \in s_t$ directly (e.g., "Is $h$ true?"), the model is prompted to generate open-ended queries $q \in z_{\text{call}}$ targeting the visual features of $\mathcal{I}$. The tool response $z_{\text{resp}}$ then becomes an independent fact extracted from $\mathcal{I}$, decoupled from the hypothesis $h$.
Formally, this modifies the dependency of the analysis step. Instead of $z_{\text{ana}}$ relying on the model's internal perception of $\mathcal{I}$ biased by $s_t$, it relies on the explicit tool output: 
% $z_{\text{resp}} = \text{Tool}(\mathcal{I}, q)$, 
\begin{equation}
    z_{\text{resp}} = \text{Tool}(\mathcal{I}, q)
\end{equation}
where $q$ is independent of $h$'s truth value.
The verification $v_t$ is then derived by comparing the content of $s_t$ against the grounded evidence $z_{\text{resp}}$, significantly reducing visual hallucinations.
The implementation details of the \texttt{ask\_questions} tool are provided in \textbf{Appendix \ref{app:tool_definitions}}.

\textbf{Example Trajectory:}
We provide an example of tool invoking in step verification in Figure \ref{fig:example}.

This decomposition decouples the perception task from the reasoning task, significantly reducing visual hallucinations.

\subsection{Sample Upweighting}

Analysis of our generated training data revealed a significant class imbalance in the set of labels $\mathcal{V}$, dominated by $v_t = \text{Correct}$. To prevent the TIM-PRM from converging to a trivial solution (predicting "Correct" regardless of $\tau_t$), we implement a sample upweighting strategy.

Let $\mathcal{D}$ be the collected dataset of trajectories. We partition $\mathcal{D}$ into two subsets: $\mathcal{D}_{+}$ containing trajectories where $\forall t, v_t \in \{\text{Correct, Neutral}\}$, and $\mathcal{D}_{-}$ containing trajectories where $\exists t, v_t = \text{Incorrect}$.
We define the training objective as a weighted negative log-likelihood loss:
\begin{equation}
    \mathcal{L}(\theta) = -\mathbb{E}_{\tau \in \mathcal{D}_{+}} [\log P_\theta(\tau)] - w \cdot \mathbb{E}_{\tau \in \mathcal{D}_{-}} [\log P_\theta(\tau)]
\end{equation}
where $w > 1$ is a scalar hyperparameter. By setting $w > 1$, we force the model to focus on the critical task of error detection and critique generation within $\tau_t$, counteracting the inherent scarcity of negative samples in mathematical reasoning data.

\section{Experiments}

\begin{table*}[htbp]
    \centering
    \caption{Performance comparison of various models on macro F1 score. The results of models marked with $^{\dagger}$ are cited from \cite{wang_visualprm_2025}}
    \label{tab:f1_score}
    % \vspace{0.2cm} % Add a little space between caption and table
    % \renewcommand{\arraystretch}{1.0} % Increase row height slightly for readability
    % \setlength{\tabcolsep}{4pt} % Adjust column spacing slightly for more data
    \small
    % \resizebox{1.0\textwidth}{!}{ % Scales table to 90% of text width
    \begin{tabular}{lcccccc}
        \toprule
        \textbf{Model} & \textbf{MMMU} & \textbf{MathVision} & \textbf{MathVerse-VO} & \textbf{DynaMath} & \textbf{WeMath} & \textbf{Overall} \\
        % \midrule
        % Random Guessing & 50.0 & 50.0 & 50.0 & 50.0 & 50.0 & 50.0 \\
        \midrule
        \multicolumn{7}{c}{\textit{Proprietary Models}} \\
        \midrule
        GPT-4o-Mini$^{\dagger}$ & 53.6 & 58.9 & 57.1 & 56.7 & 58.5 & 57.9 \\
        GPT-4o$^{\dagger}$ & 56.3 & 60.2 & 59.7 & 59.0 & 63.3 & 60.3 \\
        Gemini-2.0-Flash$^{\dagger}$ & 58.5 & 60.1 & 62.8 & 66.7 & 58.7 & 62.3 \\
        \midrule
        \multicolumn{7}{c}{\textit{Open-source Models}} \\
        \midrule
        MiniCPM-V2.6-8B$^{\dagger}$  & 44.9 & 50.9 & 58.9 & 46.7 & 57.4 & 50.4 \\
        LLaVA-OV-7B$^{\dagger}$  & 45.7 & 43.0 & 42.2 & 44.7 & 52.5 & 44.4 \\
        LLaVA-OV-72B$^{\dagger}$  & 46.1 & 48.4 & 53.0 & 57.0 & 57.3 & 52.3 \\
        Qwen2.5-VL-7B$^{\dagger}$  & 53.1 & 51.8 & 47.8 & 51.3 & 54.2 & 51.0 \\
        Qwen2.5-VL-72B$^{\dagger}$  & \textbf{59.2} & 59.0 & 59.7 & 62.9 & 62.3 & 60.5 \\
        InternVL2.5-8B$^{\dagger}$ & 47.1 & 45.5 & 47.8 & 50.3 & 50.8 & 48.0 \\
        InternVL2.5-26B$^{\dagger}$ & 48.8 & 47.4 & 49.2 & 50.4 & 51.4 & 49.2 \\
        InternVL2.5-38B$^{\dagger}$ & 51.5 & 48.4 & 50.9 & 51.8 & 52.5 & 50.8 \\
        InternVL2.5-78B$^{\dagger}$ & 52.0 & 51.7 & 53.7 & 50.8 & 52.5 & 52.6 \\
        Qwen3-VL-2B & 50.4 & 50.9 & 49.6 & 48.7 & 58.3 & 51.1 \\
        Qwen3-VL-8B & 56.6 & \textbf{61.7} & 59.9 & 61.4 & 62.7 & 61.1 \\
        VisualPRM-8B & 54.9 & 56.1 & 53.0 & 57.5 & 55.1 & 55.9 \\
        MM-PRM-8B & 51.2 & 55.4 & 54.9 & 58.1 & 56.5 & 55.5 \\
        \midrule
        \rowcolor{highlightgray}TIM-PRM-2B & 55.8 & 57.6 & 61.7 & 64.3 & 58.0 & 60.3 \\
        \rowcolor{highlightgray}TIM-PRM-8B & {58.3} & {58.3} & \textbf{61.9} & \textbf{65.9} & \textbf{63.9} & \textbf{61.7} \\
        \bottomrule
    \end{tabular}
    % }
\end{table*}

\subsection{Experiment Setup}

\paragraph{Benchmarks.} To evaluate the efficacy of TIM-PRM in verifying multimodal mathematical reasoning, we employ VisualProcessbench as the main benchmark, which contains reasoning trajectories from a diverse set of policy models on 5 benchmarks with human-annotated step-level correctness. Specifically, we report the subset results on \textbf{MMMU}~\cite{yue_mmmu_2024}, \textbf{MathVision}~\cite{wang_measuring_2024}, \textbf{MathVerse-VO}~\cite{zhang_mathverse_2024}, \textbf{DynaMath}~\cite{zou_dynamath_2024}, and \textbf{WeMath}~\cite{qiao_we-math_2025}. These datasets cover a wide spectrum of visual reasoning tasks, ranging from geometry and function analysis to complex multi-step logical deduction. We specifically focus on the subsets of these benchmarks that require step-by-step reasoning, adhering to the standard evaluation protocols defined in previous process supervision literature. We consider the neutral steps as a correct step following \citet{lightman_lets_2023}.

\paragraph{Baselines.} We compare TIM-PRM against a comprehensive suite of baselines, categorized into three distinct groups:
\begin{itemize}
    \item \textbf{Proprietary MLLMs:} We evaluate state-of-the-art closed-source models including \textbf{GPT-4o} \cite{noauthor_gpt-4o_nodate}, \textbf{GPT-4o-Mini} \cite{noauthor_gpt-4o_nodate}, and \textbf{Gemini-2.0-Flash} \cite{team_gemini_2024}. These models are prompted to act as verifiers (MLLM-as-a-PRM) using zero-shot instructions to critique the correctness of reasoning steps \cite{wang_visualprm_2025}.
    \item \textbf{Open-source MLLMs:} We include leading open-weights models such as \textbf{MiniCPM-V2.6} \cite{yao_minicpm-v_2024}, \textbf{LLaVA-OV} \cite{li_llava-onevision_2024}, \textbf{Qwen2.5-VL} \cite{bai_qwen25-vl_2025}, \textbf{InternVL2.5} \cite{chen_expanding_2025}, and \textbf{Qwen3-VL} \cite{bai_qwen3-vl_2025} across various sizes ranging from 7B to 78B. Similar to the proprietary models, these are evaluated in a generative verifier setting.
    \item \textbf{Specialized Multimodal PRMs:} We compare against \textbf{VisualPRM-8B} \cite{wang_visualprm_2025} and \textbf{MM-PRM-8B} \cite{du_mm-prm_2025}, which are trained specifically for process reward modeling using traditional outcome-based supervision or standard preference optimization techniques without tool integration.
\end{itemize}

\paragraph{Training data.} We use the MathV360K subset of the VisualPRM400K dataset because it ensembles a variety of datasets, containing about 20.1k samples. After format filtering and consensus filtering \cite{zhang2025lessonsdevelopingprocessreward} with MLLM-as-a-PRM implemented by Qwen3-VL-30B-A3B-Instruct, about 13k samples with reliable process labels remains, with a pass rate of 65\%.

\paragraph{Implementation Details.} Our TIM-PRM models are initialized from the Qwen3-VL-2B-Instruct and Qwen3-VL-8B-Instruct checkpoints. We employ Low-Rank Adaptation (LoRA) for parameter-efficient fine-tuning, setting the LoRA rank to 32 and alpha to 64. The models are trained for 3 epochs with an effective batch size of 64. We use the AdamW optimizer with a peak learning rate of $1e^{-4}$ and a cosine learning rate scheduler with a 0.1 warmup ratio. For the tool execution environment, unless otherwise specified, the external reasoning agent is powered by Qwen3-VL-30B-A3B-Instruct to ensure high-fidelity tool responses. All experiments are conducted on a server with 8 NVIDIA A800 GPUs.

\begin{table*}[htbp]
    \centering
    \caption{Analysis of First Incorrect Step Identification (FISI) F1 Score across different benchmarks.}
    \label{tab:fisi}
    % \vspace{0.2cm}
    % \renewcommand{\arraystretch}{1.1}
    % \setlength{\tabcolsep}{3pt} % Removed tight spacing as table is now narrower
    \small 

    \begin{tabular}{lcccccc}
        \toprule
        \textbf{Model} & \textbf{MMMU} & \textbf{MathVision} & \textbf{MathVerse-VO} & \textbf{DynaMath} & \textbf{WeMath} & \textbf{Overall} \\
        \midrule
        VisualPRM-8B & \textbf{16.4} & 10.2 & 9.4 & 8.9 & 6.4 & 9.9 \\
        MM-PRM-8B & 6.4 & 15.2 & 15.3 & 16.4 & 12.2 & 14.4 \\
        Qwen3-VL-2B & 4.6 & 7.9 & 5.1 & 4.4 & 7.8 & 6.0 \\
        Qwen3-VL-8B & 10.6 & 18.2 & 16.9 & 18.4 & 18.8 & 17.2 \\
        \midrule
        \rowcolor{highlightgray}TIM-PRM-2B & 12.7 & 22.4 & 25.8 & 25.1 & 23.9 & 23.4 \\
        \rowcolor{highlightgray}TIM-PRM-8B & 13.4 & \textbf{26.2} & \textbf{29.6} & \textbf{26.7} & \textbf{24.9} & \textbf{26.4} \\
        \bottomrule
    \end{tabular}
\end{table*}

\begin{table*}[htbp]
    \centering
    \caption{The performance of TIM-PRM with Independent Question Asking tool implemented with different models.}
    \label{tab:tool_strength}
    \small
    % \vspace{0.2cm} % Add a little space between caption and table
    % \renewcommand{\arraystretch}{1.1} % Increase row height slightly for readability
    \setlength{\tabcolsep}{4pt} % Adjust column spacing slightly for more data

    \begin{tabular}{lcccccc}
        \toprule
        \textbf{Tool} & \textbf{MMMU} & \textbf{MathVision} & \textbf{MathVerse-VO} & \textbf{DynaMath} & \textbf{WeMath} & \textbf{Overall} \\
        \midrule
        Qwen3-VL-2B & 52.8 & 57.1 & 60.2 & 60.9 & 56.5 & 58.6 \\
        Qwen3-VL-8B & 57.0 & 59.4 & 61.4 & 63.2 & 57.3 & 60.7 \\
        Qwen3-VL-30B & 55.8 & 57.6 & 61.7 & 64.3 & 58.0 & 60.3 \\
        \bottomrule
    \end{tabular}
\end{table*}

\begin{table*}[htbp]
    \centering
    \caption{The effectiveness of upweighting incorrect samples during training.}
    \label{tab:upweighting}
    % \vspace{0.2cm} % Add a little space between caption and table
    \renewcommand{\arraystretch}{1.1} % Increase row height slightly for readability
    \setlength{\tabcolsep}{4pt} % Adjust column spacing slightly for more data
    \small

    \begin{tabular}{lcccccc}
        \toprule
        \textbf{Upweighting Factor} & \textbf{MMMU} & \textbf{MathVision} & \textbf{MathVerse-VO} & \textbf{DynaMath} & \textbf{WeMath} & \textbf{Overall} \\
        \midrule
        $w=1$ & 51.1 & 53.7 & 58.8 & 59.0 & 57.0 & 56.7 \\
        $w=2$ & 52.0 & 56.6 & 60.3 & 58.9 & 59.5 & 58.4 \\
        $w=4$ & 59.1 & 54.1 & 60.2 & 60.9 & 63.6 & 59.1 \\
        $w=8$ & 53.1 & 58.2 & 62.1 & 61.7 & 58.7 & 60.0 \\
        $w=10$ & 55.8 & 57.6 & 61.7 & 64.3 & 58.0 & 60.3 \\
        \bottomrule
    \end{tabular}
\end{table*}

\subsection{Main results}

\paragraph{Step-wise Verification Performance.} 
We present the step-wise verification results in Table \ref{tab:f1_score}. Our proposed TIM-PRM-8B demonstrates exceptional performance, achieving an overall F1 score of \textbf{61.7}, which establishes a new state-of-the-art among open-weight models. Notably, TIM-PRM-8B significantly outperforms significantly larger models, including Qwen2.5-VL-72B (60.5) and InternVL2.5-78B (52.6), despite having an order of magnitude fewer parameters. This result underscores the efficiency of our tool-integrated framework, which effectively leverages external verification to compensate for parameter limitations. Even our smaller variant, TIM-PRM-2B, remains highly competitive, surpassing the 7B-scale baselines (e.g., LLaVA-OV-7B, Qwen2.5-VL-7B) and outperforming dedicated scalar PRMs like VisualPRM-8B by a clear margin (60.3 vs. 55.9). These findings validate that explicitly decoupling perception via tool use is a more effective strategy for verification than simply scaling model size or relying on implicit scalar rewards.

\paragraph{First Incorrect Step Identification.} 
While step-wise accuracy is informative, the ability to pinpoint the \textit{exact} location where reasoning diverges is critical for efficient search pruning. We report the First Incorrect Step Identification (FISI) F1 scores in Table \ref{tab:fisi}. Here, the advantage of TIM-PRM is even more pronounced. TIM-PRM-8B achieves an overall FISI score of \textbf{26.4}, improving over the scalar baseline VisualPRM-8B (9.9) by nearly \textbf{165\%}. 
Standard discriminative PRMs often struggle with the "needle in a haystack" problem—failing to distinguish the first subtle visual hallucination from subsequent consequent errors. By actively querying the visual environment, TIM-PRM can locally ground each step, leading to far more precise error localization. This capability is pivotal for downstream applications like Tree-of-Thoughts or Best-of-N search, where early pruning of incorrect branches saves substantial compute.

\subsection{Ablation Studies}

\textbf{Impact of Tool Strength.} 
As shown in Table \ref{tab:tool_strength}, the efficacy of our framework relies on the reliability of the external feedback. In Table 3, we analyze how the capability of the model used for the `ask\_questions` tool affects overall verification performance. We observe a consistent scaling law: replacing the tool backend from Qwen3-VL-2B to Qwen3-VL-30B improves the verifier's overall F1 score from 58.6 to 60.3. This suggests that the verifier effectively offloads the burden of fine-grained visual perception to the tool. As the tool becomes more capable of answering open-ended visual queries accurately, the verifier's judgments become more grounded, reducing false positives caused by visual hallucinations.

\begin{table*}[htbp]
    \centering
    \caption{Tool-calling frequency of correct and incorrect steps. \textbf{Cor} represents correct step and \textbf{Inc} represents Incorrect step.}
    \label{tab:tool_freq}
    % \vspace{0.2cm}
    \renewcommand{\arraystretch}{1.1}
    \setlength{\tabcolsep}{3pt} % Tighten column spacing
    \small % Slightly smaller text to fit the table width

    \begin{tabular}{lcccccccccccc}
        \toprule
        {\textbf{Model}} & \multicolumn{2}{c}{\textbf{MMMU}} & \multicolumn{2}{c}{\textbf{MathVision}} & \multicolumn{2}{c}{\textbf{MathVerse-VO}} & \multicolumn{2}{c}{\textbf{DynaMath}} & \multicolumn{2}{c}{\textbf{WeMath}} & \multicolumn{2}{c}{\textbf{Overall}} \\
        \cmidrule(lr){2-3} \cmidrule(lr){4-5} \cmidrule(lr){6-7} \cmidrule(lr){8-9} \cmidrule(lr){10-11} \cmidrule(lr){12-13}
         & \textbf{Cor} & \textbf{Inc} & \textbf{Cor} & \textbf{Inc} & \textbf{Cor} & \textbf{Inc} & \textbf{Cor} & \textbf{Inc} & \textbf{Cor} & \textbf{Inc} & \textbf{Cor} & \textbf{Inc} \\
        \midrule
        TIM-PRM-2B & 26.4 & 20.9 & 28.8 & 26.9 & 23.6 & 23.5 & 22.9 & 24.3 & 24.3 & 21.5 & 25.0 & 24.1 \\
        TIM-PRM-8B & 22.4 & 17.0 & 25.1 & 22.4 & 21.1 & 20.4 & 19.8 & 18.5 & 18.1 & 19.6 & 21.6 & 20.4 \\
        \bottomrule
    \end{tabular}
\end{table*}

\textbf{Effectiveness of Sample Upweighting.} 
Table \ref{tab:upweighting} highlights the critical importance of our training strategy. Training without upweighting ($w=1$) yields a baseline score of 56.7. As we increase the sampling weight of incorrect trajectories, forcing the model to attend more to error cases, performance steadily improves, peaking at 60.3 with $w=10$. This confirms our hypothesis that without intervention, generative verifiers tend to collapse into a mode of "lazy agreement" (sycophancy), dominated by the prevalence of correct steps in the training corpora. Upweighting effectively counteracts this class imbalance, sharpening the model's critical faculties.

\begin{figure}[t]
    \centering
    % Placeholder for performance comparison and weight function plots
    \begin{subfigure}[b]{0.48\textwidth}
        \centering
        % Placeholder for performance plot
        \includegraphics[width=\linewidth]{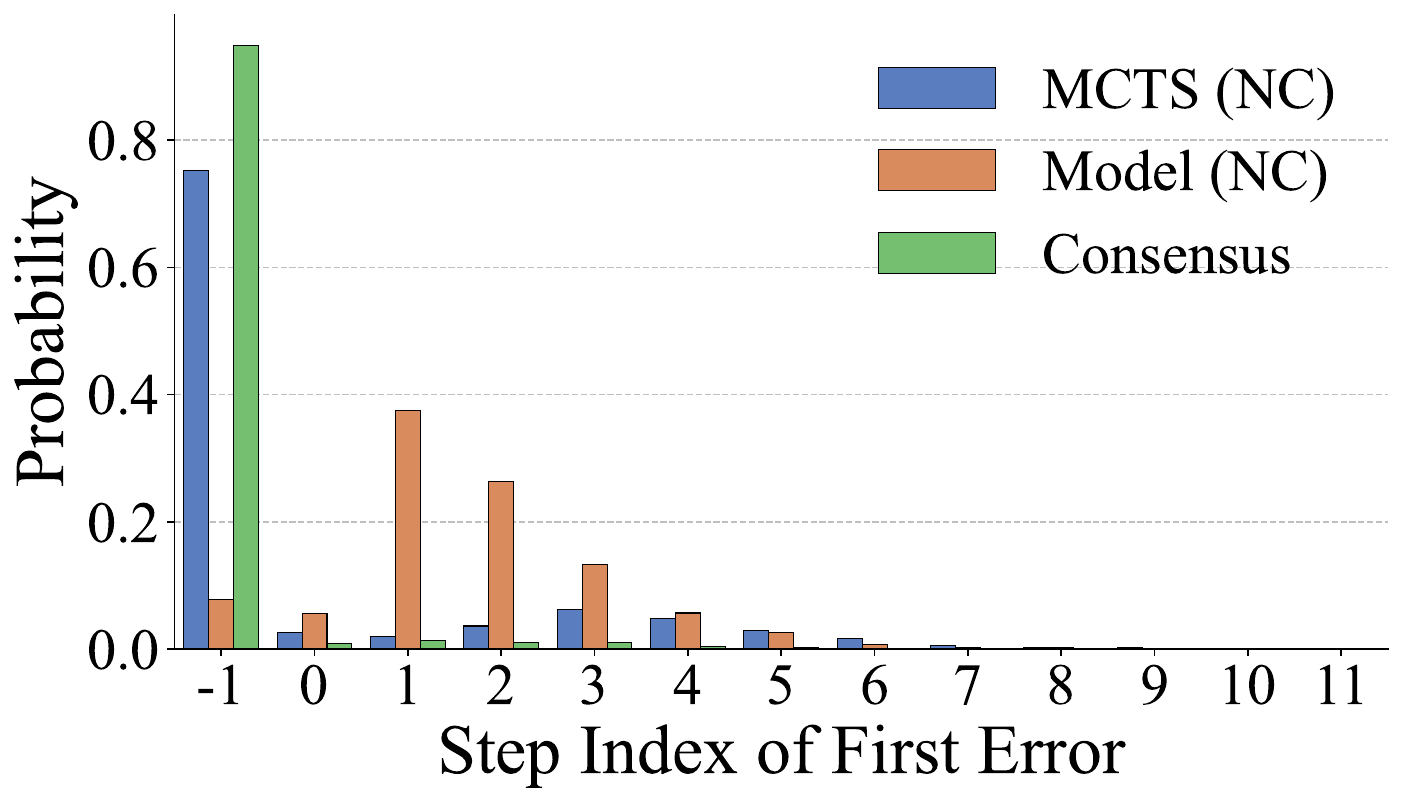}
        \caption{Analysis of First Error Identification.}
        % \vspace{3cm}
        % \caption{Performance on MATH dataset.}
        % \label{fig:MATH_scaling}
    \end{subfigure}
    % \hfill
    \begin{subfigure}[b]{0.48\textwidth}
        \centering
        \includegraphics[width=\linewidth]{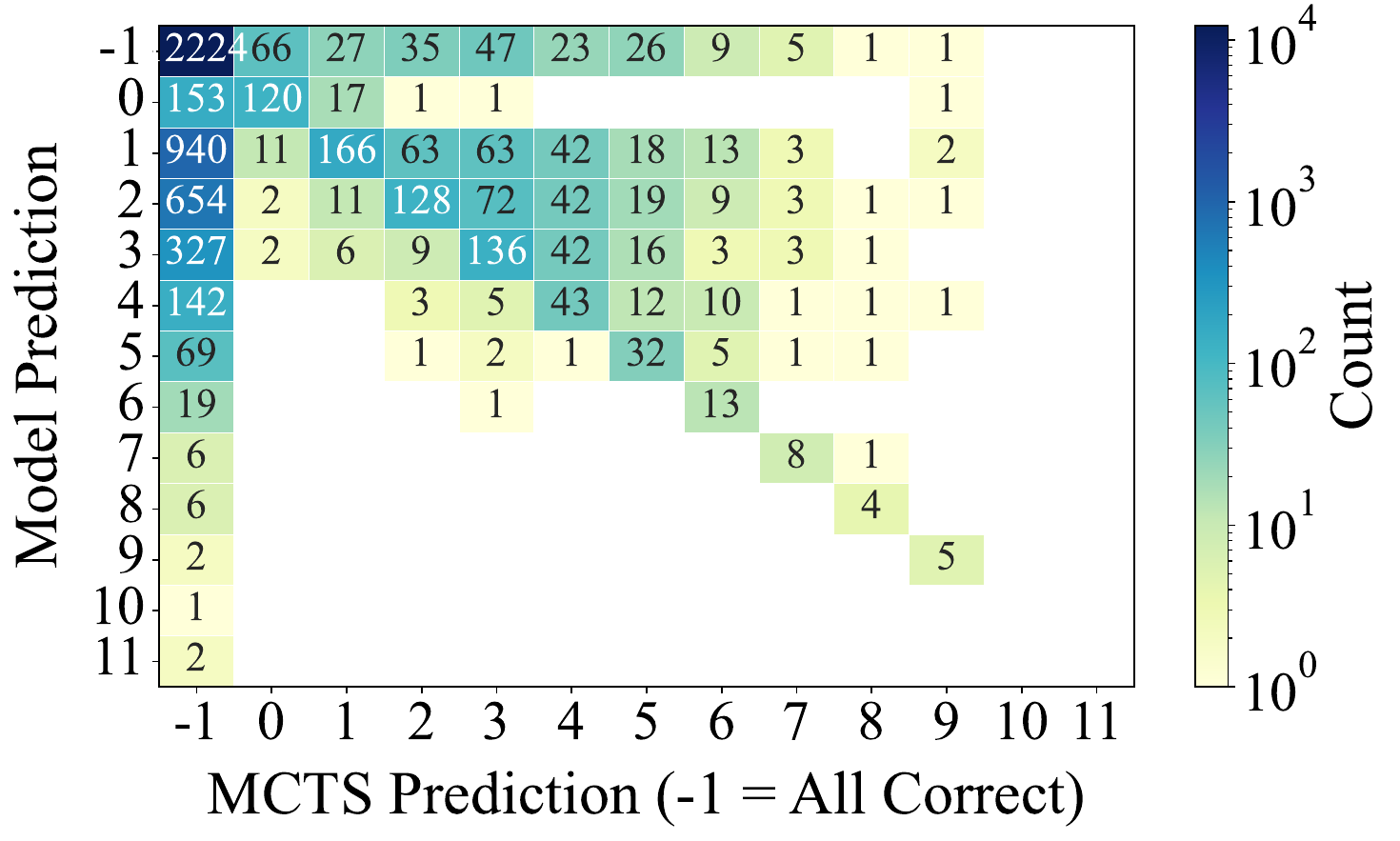}
        \caption{Analysis of Data Filtering Consistency.}
        % \vspace{3cm}
        % \caption{Performance of MATH500 dataset.}
        % \label{fig:MATH500_scaling}
    \end{subfigure}
    \caption{Training data analysis.}
    \label{fig:data_distr}
\end{figure}

\textbf{Tool Invocation Analysis.}
A potential concern with agentic frameworks is that models might overuse tools for trivial steps or underuse them for complex ones. Table \ref{tab:tool_freq} presents the frequency of tool calls across correct and incorrect steps. We observe a balanced distribution: for TIM-PRM-8B, tools are invoked in roughly 21.6\% of correct steps and 20.4\% of incorrect steps. This consistency indicates that the model has learned to trigger tools based on the \textit{nature} of the task (e.g., needing to read a chart) rather than the latent correctness of the text. 
% The slightly higher usage in correct steps may reflect the successful verification of complex visual facts that would otherwise be hallucinated without tool support.

\textbf{Data Filtering Analysis.} 
To validate the quality of our training data construction pipeline, we analyze the consistency between the initial MCTS signals and our teacher model's judgments. Figure \ref{fig:data_distr} illustrates this relationship. 
The heatmap (Figure \ref{fig:data_distr}b) visualizes the confusion matrix between the raw MCTS outcomes (x-axis) and the model's predicted first error step (y-axis). While there is a strong diagonal correlation, indicating general agreement, the significant off-diagonal density reveals the noise inherent in outcome-based MCTS labels. Many trajectories labeled as "correct" by MCTS (column -1) are identified as containing errors by the model, highlighting "false positive" reasoning paths that arrived at the correct answer serendipitously.
Figure \ref{fig:data_distr}a further plots the distribution of the first error index. The "Consensus" distribution represents our final filtered dataset. By requiring agreement between high-confidence MCTS rollouts and the teacher's explicit verification, we filter out the noise observed in the raw distributions. The high density at index -1 for the Consensus data confirms that our filtering pipeline successfully curates a high-quality dataset of fully correct reasoning chains, while accurately preserving distinct error locations for training the critic.

\section{Conclusion}

In this paper, we introduced \textbf{TIM-PRM}, a novel agentic verification framework that fundamentally shifts the paradigm of process supervision from passive scoring to active, tool-integrated investigation. Addressing the critical shortcomings of existing methods—namely, scalar rewards and the prevalence of sycophantic hallucinations—we proposed a structured approach where the verifier explicitly plans, queries evidence via independent question asking, and analyzes results before rendering a verdict. This methodology effectively decouples visual perception from the reasoning context, allowing for grounded verification that is resilient to confirmation bias.
Our extensive empirical evaluation demonstrates the superiority of this approach. The 8B parameter TIM-PRM outperforms significantly larger models such as Qwen2.5-VL-72B and InternVL2.5-78B in step-wise verification accuracy and the First Incorrect Step Identification metric, surpassing traditional scalar PRMs by a wide margin.
Future work will explore extending this agentic verification paradigm to broader domains beyond mathematics and investigating more complex tool-use strategies to further enhance the interpretability and reliability of multimodal reasoning.

\section*{Limitations}

While TIM-PRM demonstrates significant improvements in verifying multimodal reasoning, there are avenues for further optimization. 
First, as an agentic framework involving explicit planning and tool interaction, our approach inherently incurs a higher inference cost compared to passive scalar verifiers. Future work could address this by exploring knowledge distillation techniques to transfer the verification capabilities into a smaller model, or by implementing adaptive verification strategies that only trigger the agentic workflow for highly uncertain steps. 
Second, our current implementation focuses primarily on visual verification via Independent Question Asking. While effective for perception, this does not fully address complex logical or calculation errors; integrating code interpreters or symbolic solvers could further enhance logical rigor. 
Finally, while we validate our method on diverse mathematical benchmarks, exploring its application to broader multimodal domains, such as general visual scene understanding or document analysis, remains a critical direction for future research.

% Reference
% For natbib users:
\bibliography{citations}
\appendix

\section{Appendix}

\subsection{Tool Definitions}
\label{app:tool_definitions}

The prompt in Figure \ref{fig:prompt} defines the `ask\_questions` tool used by the model. It is designed to query an external MLLM to gather visual information without the bias of the solution text.

\begin{tcolorbox}[
    enhanced,
    breakable,
    colback=promptBody,
    colframe=promptHeader,
    colbacktitle=promptHeader,
    title=\textbf{Verification Trajectory with Tools},
    fonttitle=\large\bfseries,
    % sharp corners,
    boxrule=1pt,
    titlerule=0mm,
    left=5mm, right=5mm, top=5mm, bottom=5mm,
    parbox=false
]
\label{fig:example}

\texttt{\#\#\# Paragraph 2}

% \vspace{0.5em}
\textbf{<planning>}
This paragraph calculates the total sales for Product 1 by summing the values from each store. I need to verify the values from the heatmap and the addition. I will use the ask\_questions tool to confirm the values in the heatmap for Product 1 across all stores.
\textbf{</planning>}

% \vspace{0.5em}
\textbf{<tool\_call>}
\{"name": "ask\_questions", "arguments": \{"target\_image": 1, "questions": ["What are the sales values for Product 1 in Store A, Store B, Store C, and Store D?"]\}\}
\textbf{</tool\_call>}

% \vspace{0.5em}
\textbf{<tool>}
The sales values for Product 1 in Store A, Store B, Store C, and Store D are 119, 177, 116, and 159, respectively.
\textbf{</tool>}

% \vspace{0.5em}
\textbf{<analyze>}
The values for Product 1 are confirmed as 119 (Store A), 177 (Store B), 116 (Store C), and 159 (Store D). The sum is 119 + 177 + 116 + 159 = 571, which matches the calculation in the paragraph. The paragraph is correct.
\textbf{</analyze>}

% \vspace{0.5em}
\textbf{<verify>}
CORRECT
\textbf{</verify>}

\end{tcolorbox}

\subsection{Prompt Structure for Trajectory Generation}
\label{app:prompt_structure}

We define the prompt for generating tool-calling trajectories in Figure \ref{fig:prompt}. This structure ensures the model adheres to the \texttt{<planning>} $\to$ \texttt{<tool\_call>} $\to$ \texttt{<analyze>} $\to$ \texttt{<verify>} workflow.
\raggedbottom

\begin{figure*}[t] % The * tells LaTeX to span both columns
    \centering
    \begin{tcolorbox}[
        enhanced,
        width=\textwidth,        % Ensure box fills the whole page width
        halign=flush left,       % Fixes body text stretching
        halign title=flush left, % Fixes title stretching
        colback=promptBody,      
        colframe=promptHeader,   
        colbacktitle=promptHeader, 
        title=\textbf{Prompt for Tool-Integrated Verification}, 
        fonttitle=\large\bfseries,
        boxrule=1pt,
        titlerule=0mm,
        left=5mm, right=5mm, top=5mm, bottom=5mm,
        parbox=false             
    ]

% --- SYSTEM SECTION ---
\textbf{[System]:}\\
You are a math teacher. Your task is to review and critique the paragraphs in solution step by step. You have access to tools to help you gather information. Use them when necessary.

\vspace{1em}
\textbf{[Available Tools]}

Function: ask\_questions\\
Description: Asks one or more questions about a specific image to gather more information. Use it when you are unsure about what you see and need confirmation. The model `\textit{<model\_name>}' will be used to answer.\\
Arguments:
\begin{itemize}[noitemsep, topsep=0pt, leftmargin=1.5em]
    \item target\_image: (Integer) The 1-based index of the image.
    \item questions: (List[String]) A list of questions to ask.
\end{itemize}
Example Usage:
\begin{verbatim}
<tool_call>{"name": "ask_questions", "arguments": 
{"target_image": 1, "questions": ["Question 1?", "Question 2?"]}}</tool_call>
\end{verbatim}
% --- USER SECTION ---
\textbf{[User]:}\\
The following is the multi-modal math problem and a solution (split into paragraphs).

\vspace{1em}
\textbf{[Math Problem]}

\{problem\_text\}

\textbf{[Solution]}

<paragraph\_1>\\
\{step\_content\_1\}\\
</paragraph\_1>

...

<paragraph\_n>\\
\{step\_content\_n\}\\
</paragraph\_n>

\vspace{1em}
Your task is to verify the correctness of each paragraph in the solution. Split your verification by \texttt{\#\#\# Paragraph \{\{ID\}\}}.

For each paragraph, you must follow this workflow:
\begin{enumerate}[leftmargin=1.5em]
    \item Start with an \texttt{<planning>} part. In this part, you should analyze whether and how tools should be called to verify the visual correctness, knowledge correctness, and logic soundness of the paragraph.
    \item Based on your planning, you can either call a tool (Step 3) or move directly to analysis (Step 4).
    \item (Optional) To call a tool, output a \texttt{<tool\_call>JSON\_BLOB</tool\_call>} section with the JSON for the tool you want to use. This will invoke the corresponding tool and return a tool response.
    \item (Mandatory) Now you have to first provide a \texttt{<analyze>} section to provide the rationale of your verification based on the tool response (if any). Then, you MUST conclude by providing a \texttt{<verify>} part with your overall judgement.
\end{enumerate}

\end{tcolorbox}
    % Optional: Add a caption if you treat this as a figure
    \caption{The system prompt used for verification.}
    \label{fig:prompt}
\end{figure*}
% --- END OF WIDE PROMPT BOX ---

\end{document}